\pdfoutput=1

\documentclass[11pt]{article}

\usepackage[]{acl}

\usepackage{times}
\usepackage{latexsym}
\newcommand{\modelname}{PLUG}
\usepackage{multirow}
\usepackage{mathtools}

\usepackage{booktabs}
\usepackage{graphicx}

\usepackage[T1]{fontenc}

\usepackage[utf8]{inputenc}

\usepackage{microtype}

\title{Knowledge-Grounded Dialogue Generation with \\ a Unified Knowledge Representation}

\author{Yu Li\thanks{~~Work was done when Yu Li was interning at MSR}~$^2$, Baolin Peng$^{1}$, Yelong Shen$^1$, Yi Mao$^1$, Lars Liden$^1$, Zhou Yu$^2$, Jianfeng Gao$^1$ \\
  $^1$Microsoft Research, Redmond, WA \\
  $^2$Columbia University, New York, NY \\
  \texttt{\{bapeng,yeshe,maoyi,laliden,jfgao\}@microsoft.com} \\
  \texttt{\{yl5016, zy2461\}@columbia.edu}
}

\begin{document}
\maketitle
\begin{abstract}
Knowledge-grounded dialogue systems are challenging to build due to the lack of training data and heterogeneous knowledge sources.
Existing systems perform poorly on unseen topics due to limited topics covered in the training data. In addition, it is challenging to generalize to the domains that require different types of knowledge sources.
To address the above challenges, we present \modelname\footnote{\textbf{P}re-trained \textbf{L}anguage model with a \textbf{U}nified knowledge representation for knowledge-\textbf{G}rounded dialogues.}, a language model that homogenizes different knowledge sources to a unified knowledge representation for knowledge-grounded dialogue generation tasks.
We first retrieve relevant information from heterogeneous knowledge sources (e.g., wiki, dictionary, or knowledge graph); Then the retrieved knowledge is transformed into text and concatenated with dialogue history to feed into the language model for generating responses. \modelname\ is pre-trained on a large-scale knowledge-grounded dialogue corpus.
The empirical evaluation on two benchmarks shows that \modelname\ generalizes well across different knowledge-grounded dialogue tasks. It achieves comparable performance with state-of-the-art methods in the fully-supervised setting and significantly outperforms other approaches in zero-shot and few-shot settings.
\end{abstract}

\section{Introduction}
Recent work has shown that conversational models can be trained in an end-to-end fashion \cite{gao2019neural, Blenderbot, DialoGPT, Meena}. Though such models can generate coherent and natural responses consistent with conversation history, there is still a clear gap between conversational AI agents and humans. The primary reason is that existing dialogue systems lack knowledge of the subject and thus cannot deep dive into specific topics with humans. In order to better incorporate knowledge into dialogue, knowledge-grounded dialogue systems have become increasingly popular.

Knowledge-grounded dialogue generation aims to generate informative and meaningful responses based on both conversation context and external knowledge sources. 
Thus far, researchers have collected knowledge-grounded dialogues for various tasks using crowdsourcing platforms, for instance, open-domain dialogues \cite{dinan2018wizard,cmu_dog_emnlp18} and conversational recommendation dialogues \cite{li2018conversational, moon2019opendialkg,hayati-etal-2020-inspired}. Workers are asked to base their replies on knowledge from structured knowledge bases \cite{moon2019opendialkg,tuan2019dykgchat} or unstructured documents \cite{dinan2018wizard,cmu_dog_emnlp18,feng2020doc2dial}. Taking advantage of recent advances in large-scale language models \cite{raffel2019exploring,lewis2020bart, guu2020realm}, researchers have also built knowledge-grounded dialogue systems by fine-tuning such language models in an end-to-end fashion \cite{shuster2021retrieval, zhao2020knowledge, li2020deux}.

\begin{table}[tb!]
\centering

\begin{tabular}{lcc}
    \toprule
    \textbf{Dataset} & \textbf{Knowledge} & \textbf{\% Topics}\\
    \midrule
    \textit{Open-domain}\\
     Wizard of Wikipedia & articles & 0.02\%\\
     CMU\_DoG & articles & 0.04\%\\
    \midrule
    \textit{Recommendation}\\
    \textsc{ReDial} & tables & 15.0\%\\
    \textsc{OpenDialKG} & graph &7.5\%\\
    
    \bottomrule
\end{tabular}
\caption{Knowledge representation and topic coverage statistics of existing knowledge-grounded dialogue datasets. \textbf{\% Topics} means the portion of topics or facts in the knowledge database covered by the dataset.}
\label{tab:datasets topics statistics}
\vspace{-3mm}
\end{table}

However, there are two critical challenges in these existing methods. First, it is expensive and time-intensive to collect knowledge-grounded dialogues. As shown in Table \ref{tab:datasets topics statistics}, most of the datasets only cover a small portion of the knowledge base. Thus, systems which only fine-tune with small training sets generalize poorly on unseen topics in the same knowledge base. Additionally, the formats of knowledge sources vary in different tasks, making the approaches unable to transfer to other domains with different knowledge sources. For example, \textsc{ReDial} \cite{li2018conversational} adopts a movie database as the knowledge source to recommend movies. Techniques on this task exploit the graph structure. It is not easy to adapt such techniques to document-grounded conversation tasks like Wizard of Wikipedia \cite{dinan2018wizard}.

In this work, we present \modelname, a model that can unify different knowledge formats for knowledge-grounded dialogue generation. First, we convert different knowledge formats (e.g., knowledge graph, knowledge base, and passages) to unstructured text, each using a different retriever. Then we use a pre-trained language model to process them into a unified representation to incorporate the knowledge into dialogue generation. We pre-train \modelname\ on different knowledge-ground dialogue corpora, including a large-scale open-domain conversation dataset from Reddit. This allows \modelname\ to learn knowledge in various formats from different tasks, and thus transfer to any knowledge-grounded dialogue task with few-shot learning techniques.

We evaluate the effectiveness of \modelname\ by applying it to an open-domain knowledge-grounded dialogue benchmark, Wizard of Wikipedia \cite{dinan2018wizard}, and a knowledge-grounded conversational recommendation benchmark, \textsc{ReDial} \cite{li2018conversational}. \modelname\ achieves results comparable to the state-of-the-art method under a fully-supervised setting.  It outperforms other methods on both tasks under zero-shot and few-shot settings, demonstrating that \modelname\ can be grounded on world knowledge in different knowledge sources and generalize to different downstream tasks.

Our contributions are three-fold: (1) We propose a novel knowledge-based pre-trained language model, \modelname, that can be applied to any knowledge-grounded dialogue tasks; (2) Our model achieves slightly better results than state-of-the-art models in fully-supervised settings and shows promising improvements over the current state-of-the-art in zero-shot and few-shot settings; (3) We present extensive experiments to explore the bottlenecks of the task and the future direction of knowledge-grounded dialogues.

\section{Approach}
\begin{figure*}
    \centering
    \includegraphics[trim={0 0 0 0},clip,width=16cm]{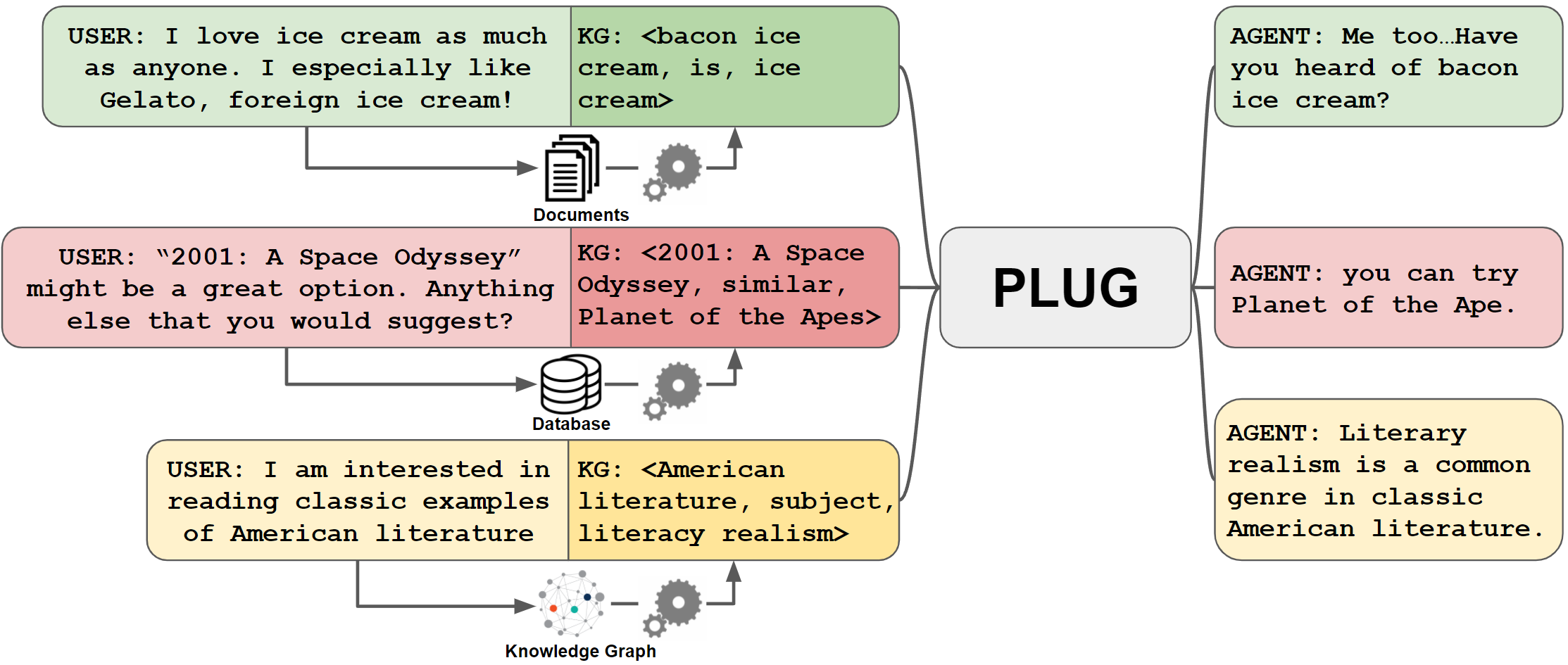}
    \caption{A diagram of \modelname. \modelname\ homogenizes different knowledge sources in different tasks to a unified knowledge representation. Then it learns to ground response generation on the unified knowledge representation.}
    \label{fig:archi}
    \vspace{-3mm}
\end{figure*}

We describe our approach in this section. Figure \ref{fig:archi} gives a diagram of our proposed method. We first introduce the background of knowledge-grounded dialogues and the backbone language model in Section \ref{method_background}. Then, we formalize the task and introduce the details of \modelname\ in Section \ref{method_archi}. Finally, we explain the training process of our \modelname, which includes the pre-training dataset selection and the data pre-processing processes in Section \ref{pre_training_datasets}.

\subsection{Background: Knowledge-Grounded Pre-training}
\label{method_background}
Traditional knowledge-grounded dialogue includes three steps: information extraction, knowledge prediction, and response generation. Previous work focuses on developing separate modules \cite{zhou2020design}. Inspired by the recent success of applying a large-scale pre-trained language model on task-oriented dialogue systems \cite{peng2020soloist, simpletod}, we explore the possibility of using a unified knowledge representation in a large-scale language model. In order to properly manage the task in a sequence-to-sequence setup, we choose T5 \cite{2020t5} as our backbone. 

T5 is a sequence-to-sequence pre-trained Transformer \cite{vaswani2017attention} model for transfer learning. It is trained by converting various language tasks into text-to-text tasks. After fine-tuning on a dialogue dataset, T5 can generate fluent and coherent responses. Nevertheless, responses are often too generic because they are not grounded on specific knowledge. \modelname\ is built on the T5 model but grounded on real-world knowledge during training, making it inherit T5's capability of producing good responses but include more knowledge.

\subsection{\modelname}
\label{method_archi}
We formulate a knowledge-grounded dialogue as:
\begin{equation}
    \mathcal{D} = \{C, R, \mathcal{S}\}
\end{equation}
where $C = \{C_i\}^{n}_{i=1}$ is a dialogue context, and $R = \{R_i\}^{n}_{i=1}$ is the response in a dialogue that has $n$ turns. $\mathcal{S}$ is the external knowledge source for task $t$. For each dialogue turn, we can formulate a knowledge-grounded dialogue generation task on a single domain $d$ as $p(R_{i}|C_{i},\mathcal{S})$.

As shown in Figure \ref{fig:archi}, each task has its own knowledge source (e.g., documents, databases, and knowledge graphs). In order to make all knowledge-grounded dialogue generation tasks able to fit in the text-to-text encoder-decoder framework, we follow T5 to feed each dialogue turn into the language model simply by concatenating the context $C_i = \{c_1, c_2, ..., c_m\}$, and essential knowledge triples $K_i = \{k_1, k_2,..., k_n\}$ as a token sequence. The essential knowledge is extracted from the knowledge source $\mathcal{S}$ and represented as text of triples. We train the model to predict the responses token sequence $R = \{r_1, r_2,...,r_k\}$. The probability of the responses is formulated as:
\begin{equation}
\begin{aligned}
    p(R_i|C_i) = \prod_{t = 1}^{k}p(r_t|C_i,K_i,r_1,...,r_{t-1})
\end{aligned}
\end{equation}
We will explain how we select and process pre-training datasets in the following sections.

\subsection{Model training process}
\label{pre_training_datasets}
We pre-trained the \modelname\ model using two datasets, Reddit Conversation \cite{galley2018end} and OpenDialKG \cite{moon2019opendialkg}. We will first present the three-step data cleaning process of Reddit Conversation in Section \ref{pre_training_datasets_reddit}, then we will introduce OpenDialKG in Section \ref{pre_training_datasets_opendialkg}. 

\subsubsection{Reddit Conversation}
\label{pre_training_datasets_reddit}
Reddit Conversation \citet{galley2018end} is a large-scale open-domain conversation dataset. It extracts the conversation threads grounded on a document from the Reddit data.\footnote{Reddit data dumps: https://files.pushshift.io/reddit/} We only keep the conversations grounded on Wikipedia passages for pre-training to recognize better the knowledge used in the dialogue. Since vanilla document-based dialogue in the dataset does not have a knowledge label for each dialogue turn, we apply a hierarchical information extraction method to obtain the essential knowledge in each turn. Our information extraction method includes three steps: knowledge retrieval, statistical ranking, and semantic ranking.

\paragraph{Knowledge Retriever.}
We use a knowledge retriever to retrieve all relevant knowledge in a single turn's response. We first extract the title of the grounding Wikipedia passage in the dialogue. Then, we retrieve knowledge triples from a large-scale knowledge graph, DBpedia \cite{lehmann2015dbpedia}. Specifically, we query the DBpedia via a public SPARQL endpoint\footnote{https://dbpedia.org/sparql} and then collect triples whose subject or object is in the Wikipedia passage in the dialogue. For example, we keep triples \textit{<Barack Obama, alma mater, Columbia University>} and \textit{<Michelle Obama, spouse, Barack Obama>} for the dialogue about Barack Obama. To carry sufficient knowledge to refine in the next step, we retrieve 500 triples for every passage.

\paragraph{Statistical Ranking.}
After retrieving adequate knowledge, we rank the corresponding triples to refine the knowledge. Specifically, we get the TF-IDF (term frequency-inverse document frequency) value for all the retrieved triples. To find the triples related to the context, we concatenate the dialogue history and the response as the query. Then we compute the cosine similarity between the query and every triple. Because every triple has the Wikipedia passage name as the subject or object, a higher cosine similarity score means the query has more similar text with the distinguished text in the triple. We rank the query-document similarity score and only keep the top-50 triples in this step. 

\paragraph{Semantic Ranking.}
The TF-IDF-based cosine similarity score only counts words overlapping between triples and the query. It will introduce triples whose overlapping words are not meaningful in the context and response. Additionally, the Reddit Conversation dataset is obtained from Reddit conversation threads. It involves many responses that are not grounded on any knowledge. In order to find the triples that have the best semantic similarity with the response and filter out ungrounded responses, in this step, we estimate the semantic similarity score with Sentence-Bert \cite{reimers2019sentence}. We rerank the 50 triples from the second step based on the score. Additionally, we abandon the dialogue turns whose best semantic similarity is lower than a threshold because the response cannot find proper knowledge, while we want to pre-train the model with knowledge-grounded turns.

\subsubsection{OpenDialKG}
\label{pre_training_datasets_opendialkg}
To generalize our model to various tasks, we also employ OpenDialKG to enrich our pre-training dataset. OpenDialKG consists of two types of tasks,  recommendations and chit-chat, across four domains. Unlike the Reddit Conversation dataset, which needs to find the knowledge grounding in every turn, the original OpenDialKG has a Knowledge graph path label for each dialogue, and a triple label for each dialogue turn. The response is grounded on the labeled triple during data collection. Thus, we use the triple in the dataset as the essential knowledge in our pre-training examples.

\section{Experiments}
We demonstrate our approach on two different downstream tasks: open-domain knowledge-grounded dialogue and conversational recommendation. Besides the fully-supervised learning setting, we also explore the performance of our approach in few-shot and zero-shot settings. We describe our implementation details in Section \ref{sec:implementation_details} in Appendix.

\subsection{Datasets and Knowledge Sources}
\label{exp_datasets}
We test our approach on Wizard of Wikipedia (WoW; \cite{dinan2018wizard}) and \textsc{ReDial}  \cite{li2018conversational}. Basic dataset statistics are listed in Table \ref{tab:dataset_statistics}.
\begin{table}[htb!]
    \centering
    \begin{tabular}{lrrr}
    \toprule
    Dataset & Train & Valid & Test \\
    \midrule
    \midrule
    \multirow{2}{*}{WoW} & \multirow{2}{*}{18,430} & Seen - 981 & 965\\
     &  & Unseen - 967 & 968\\
    \textsc{ReDial} & 8,004 & 1,001 & 1,001\\
    \bottomrule
    \end{tabular}
    \caption{Number of conversations in Wizard of Wikipedia (WoW) and \textsc{ReDial}}
    \label{tab:dataset_statistics}
    \vspace{-3mm}
\end{table}

\paragraph{Wizard of Wikipedia.} This dataset \cite{dinan2018wizard} is collected on Amazon Mechanical Turk. Each conversation happens between a ``wizard'' who has access to knowledge about a specific topic, and an ``apprentice'' who is interested in the topic. The wizard's response is grounded on a Wikipedia article in each turn. The data is split as a training set, a validation set, and a test set. The test set has two subsets: Test Seen and Test Unseen. Test Seen contains conversations whose topics are seen in the training set, while topics in Test Unseen are not seen in the training or validation set. To extract the essential knowledge in each dialogue turn, we first keep the top five passages retrieved by the TF-IDF retriever in \citet{shuster2021retrieval}. Then we use an Open Information Extraction (OpenIE) annotator\footnote{https://nlp.stanford.edu/software/openie.html} to extract the top three triples from the passages as our essential knowledge. The pre-processing is conducted with the code published on ParlAI.\footnote{https://github.com/facebookresearch/ParlAI}

\paragraph{\textsc{ReDial}.} \textsc{ReDial} \cite{li2018conversational} is also collected on Amazon Mechanical Turk. Two crowd-workers, a ``movie seeker'' and ``movie recommender,'' are randomly paired. The recommender has access to a movie database and can recommend movies based on movie information, such as actors and movie genres. There are 6,924 different movies mentioned in 51,699 movie slots in the dataset. We follow \citet{li2018conversational} to split the dataset into training, validation, and test sets. Since recommenders use movie-related knowledge when they recommend movies to seekers, we use it as the essential knowledge for a given turn in this dataset. We experiment with three knowledge sources: (1) We query the movie names mentioned in the dialogue context and retrieve similar movies from the knowledge graph \textbf{DBpedia}, mentioned in Section \ref{pre_training_datasets}, and then input the similar movies in a triple format as the essential knowledge. (2) We query the movie names mentioned in the context and retrieve movie comments from \textbf{MovieLens}.\footnote{https://grouplens.org/datasets/movielens/}, then use the keywords in the comments as the essential knowledge. (3) We use the output of the recommender module in \textbf{KGSF} \cite{zhou2020improving}, which is the state-of-the-art system on this dataset.

\subsection{Baselines}
We compare the known best models from different datasets in the following experiments. For the Wizard of Wikipedia dataset, we choose the retrieval-augmented generation (RAG) model from \citet{shuster2021retrieval}. It retrieves wiki documents and generates responses based on the documents. We compare \modelname\ with this document-based generation method to see the impact of our essential knowledge format. We choose the RAG model also using T5 as the baseline for a fair comparison.

For the \textsc{ReDial} dataset, we choose the current state-of-the-art: KBRD \cite{chen2019knowledgebased} and KGSF  \cite{zhou2020improving} as our baselines. Both use a recommender module to predict the recommendation item in the next turn and a generation model to generate the response. All baseline results are from \citet{zhou-etal-2021-crslab}. To investigate the best performance of our approach, We also use the recommender from KGSF as a knowledge source in our system and compare it with other knowledge sources we mentioned in Section \ref{exp_datasets}. As an ablation study, we also explore the performance of vanilla T5 on both tasks to see the performance gain brought by our pre-training process.

\subsection{Metrics}
For evaluation, we report the performance with standard automatic metrics: BLEU-4 (B4) \cite{papineni2002bleu}, ROUGE-L (RL) \cite{lin2004rouge}, and unigram overlap (F1) of the generated responses. Besides that, for the Wizard of Wikipedia dataset, we follow \citet{shuster2021retrieval} to report the overlapping unigrams between the model's generation and the knowledge on which the human grounded during dataset collection (KF1), attempting to capture whether a model is speaking knowledgeably. On the other hand, for the \textsc{ReDial} dataset, we follow previous work \cite{chen2019knowledgebased, zhou2020improving, wang2021finetuning} to report distinct-n (Dist-n) at the sentence level to evaluate the diversity of the model's generation. We also evaluate whether the ground truth movie recommendation can be found in the generated response and report it as the recommendation item recall in responses (Rec).


\subsection{Fully-Supervised Results}
We first evaluate \modelname\ with all training examples in the training sets to compare its performance with other state-of-the-art systems. Additionally, we experiment with using golden knowledge in the input to explore the upper bound of our method. 

Table \ref{tab:result_wow} shows the Wizard of Wikipedia Test Seen and Test Unseen results. We can see that \modelname\ with retrieved knowledge achieves better BLEU-4, ROUGE-L, and F1 scores than the RAG method and the model without adding knowledge in the input, on both seen and unseen topics. It suggests that our essential knowledge format helps the model generate responses to ground knowledge better. We also observe that \modelname\ outperforms the model without pre-training on all metrics, which means our pre-training can boost this task.

We list \textsc{ReDial}'s results in Table \ref{tab:result_redial}. We compare our approach to the state-of-the-art systems and T5-Large models without pre-training. Additionally, we include a comparison to models with different knowledge sources as described in Section \ref{exp_datasets}. It shows that our best model (\modelname $+$KGSF) achieves the new state-of-the-art results on the recommendation item recall metric and distinct metrics. This result is understandable given that our approach is built upon pre-trained language models. Similarly, we also observe noticeable performance gains for the pre-training on this task. However, compared to systems with currently available knowledge sources, it is immediately apparent that the system with golden knowledge outperforms the current state-of-the-art on all metrics by a large margin. This huge gap implies that current knowledge retrievers are the main bottleneck for the conversational recommendation task. We will discuss more details in Section \ref{discussion}.

Overall, we observe noticeable improvement brought by pre-training on both tasks, but it is less significant than expected. It implies that the knowledge grounding pattern in the response is limited; a complete training set is more than enough for the T5-Large model to learn the generation task. We will discuss more details in zero-shot and few-shot settings in the following subsections.

\begin{table*}[htb!]
    \centering
    \small
    \begin{tabular}{l|cccc|cccc}
    & \multicolumn{4}{c}{Test Seen} &  \multicolumn{4}{c}{Test Unseen}\\
    Model & BLEU4 & ROUGE-L & F1 & KF1 & BLEU4 & ROUGE-L & F1 & KF1\\
    \midrule
    \midrule
    RAG-T5-Large \cite{shuster2021retrieval}     & 3.8 &  22.1 & 21.9 & 25.9 & 2.8	& 20.4 & 20.5 & 21.9\\
    T5-Large-w/o Knowledge   & 4.1 & 18.0 & 18.3 &  19.2 & 2.1 & 15.4 & 21.4 & 13.9\\
    T5-Large-Retrieved Knowledge     &  5.8 & 21.8 &  25.8 &  22.6 & 3.4 & 19.2 & 22.7 & 17.6\\
    T5-Large-Golden Knowledge     & 11.3 & 30.8 & 35.6 & 46.8 & 8.7 & 28.4 & 33.0 & 43.6\\ 
    \midrule
    \modelname-Retrieved Knowledge     &  6.0 & 22.3 & 26.5 &  22.4 & 3.5 & 19.5 & 23.3 & 18.6\\
    \modelname-Golden Knowledge     & 11.5 & 31.1 & 36.0 & 47.8 & 8.8 & 29.0 & 33.4 & 46.0\\ 
    \end{tabular}
    \caption{Fully-supervised results on Wizard of Wikipedia Test Seen and Test Unseen Sets.}
    \label{tab:result_wow}
    \vspace{-3mm}
\end{table*}

\begin{table}[htb!]
    \centering
    \small
    \begin{tabular}{lccccc}
    Model & B4 & RL & DIST2 & DIST4 & Rec \\
    \midrule
    \midrule
    KBRD & 1.8 & 16.5 & 0.48 & 0.67 & 0.7 \\
    KGSF & 2.3 & 13.1 & 0.49 & 1.28 & 0.9 \\
    \midrule
    \multicolumn{6}{l}{\textbf{T5-Large}}\\
    +w/o KG & 3.7 & 18.3 & 0.72 & 1.10 & 3.4\\
    +Golden & 10.4 & 32.7 & 1.17 & 1.60 & 83.5\\
    +KGSF & 3.7 & 17.4 & 1.13 & 2.02 & 4.7\\
    \midrule
    \multicolumn{6}{l}{\textbf{\modelname}}\\
    +w/o KG & 3.9 & 19.6 & 0.78 & 1.31 & 3.7\\
    +Golden & 10.6 & 33.5 & 1.26 & 1.81 & 84.3\\
    +DBpedia & 3.3 & 18.3 & 0.45 & 0.66 & 0.8\\
    +MovieLens & 3.4 & 17.8 & 0.91 & 1.34 & 2.4\\
    +KGSF & 3.8 & 18.0 & 1.51 & 2.84 & 5.3\\
    \end{tabular}
    \caption{Fully-supervised results on \textsc{ReDial}.}
    \label{tab:result_redial}
    \vspace{-3mm}
\end{table}

\subsection{Zero-Shot and Few-Shot Results}
\begin{figure*}
    \centering
    \includegraphics[trim={0 0 0 0},clip,width=16cm]{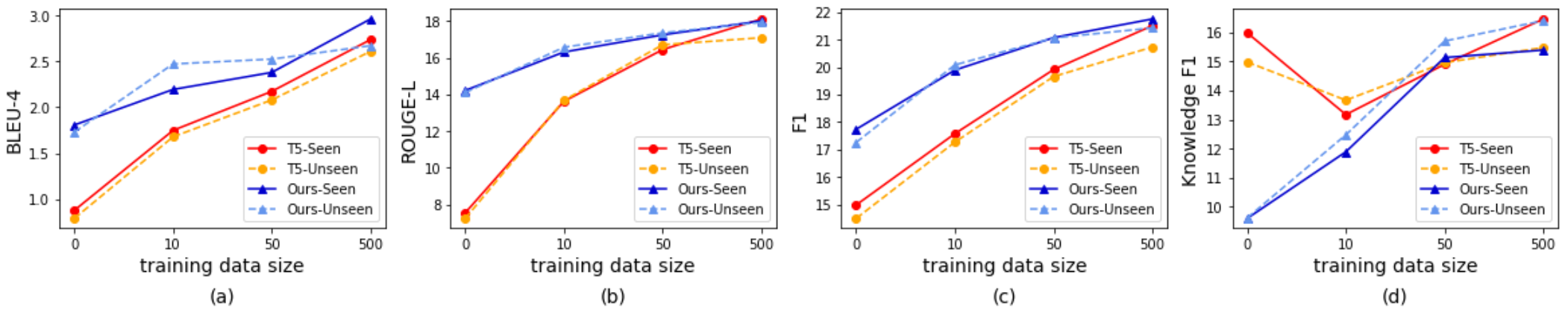}
    \caption{Zero-shot and few-shot results on Wizard of Wikipedia Test Seen and Test Unseen sets.}
    \label{fig:few_shot_wow}
    \vspace{-3mm}
\end{figure*}

\begin{figure*}
    \centering
    \includegraphics[trim={0 0 0 0},clip,width=16cm]{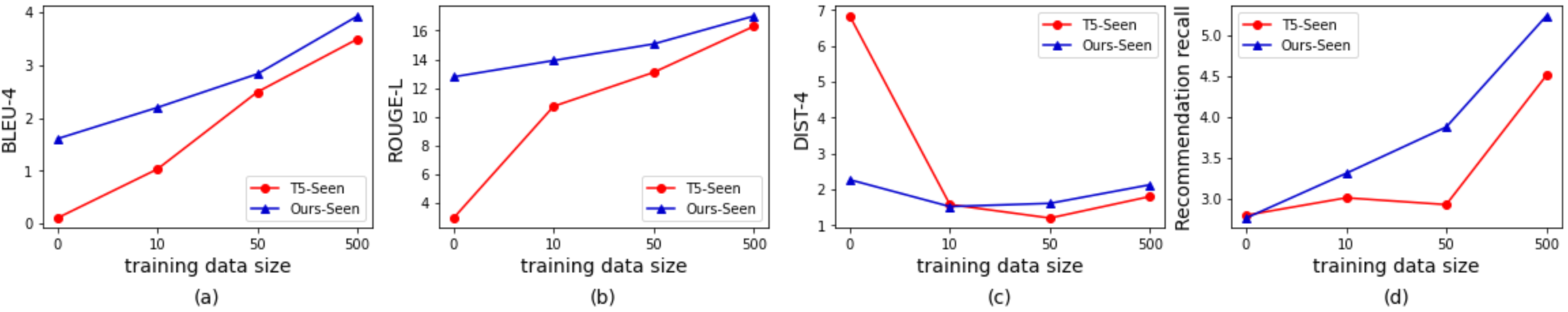}
    \caption{Zero-shot and few-shot results on \textsc{ReDial}.}
    \label{fig:few_shot_redial}
    \vspace{-3mm}
\end{figure*}

We focus on zero-shot and few-shot settings because it is more realistic to evaluate dialogue systems. Specifically, we randomly sample 10/50/500 dialogues with different topics from the training sets and observe performance on the complete test sets. We also evaluate under a zero-shot setting. We experiment with knowledge retrieved by existing retrievers on both tasks to set a realistic setting. We compare our models to those without pre-training to explore how our pre-training benefits the model's few-shot learning capability. Wizard of Wikipedia's experiments results are in Figure \ref{fig:few_shot_wow}, and \textsc{ReDial}'s results are in Figure \ref{fig:few_shot_redial}. Note that for Wizard of Wikipedia, topics in original Test Seen set may not be seen during training in this setting since we only use a small portion of data in the original training set. We use original Test Seen and Test Unseen sets to compare with fully-supervised results. As can be seen in Figure \ref{fig:few_shot_wow} (a)-(c), \ref{fig:few_shot_redial} (a)-(b), \modelname\ maintains a higher BLEU-4, ROUGE-L, and F1 scores on both tasks when training with less than 500 dialogues. It means \modelname\ obtains knowledge-grounded generation ability from pre-training and can generalize to different tasks. 

Figure \ref{fig:few_shot_wow} (d) shows that models without pre-training achieve a higher knowledge F1 score under a zero-shot setting for the Wizard of Wikipedia dataset. In contrast, it achieves a deficient performance on the language quality-related metrics, which implies that models only copy knowledge words but generate gibberish responses without training. Nevertheless, \modelname\ still generates knowledge-grounded responses with a lower knowledge F1 score out-of-the-box. This result also suggests that we should only consider knowledge F1 scores when the model has decent scores on language quality metrics.

For the \textsc{ReDial} dataset, Figure \ref{fig:few_shot_redial} (d) shows that there is not as much improvement in recommendation item recall brought by pre-training when compared to BLEU-4 and ROUGE-L on a zero-shot setting. However, we observe a noticeable difference between \modelname\ and the T5 model, which means \modelname\ learns to generate with grounded knowledge faster than the T5 model. The unusually high DIST-4 of T5 in Figure \ref{fig:few_shot_redial} (d) is caused by diverse but irrelevant responses. It is also demonstrated by low BLEU-4 and ROUGE-L scores in Figure \ref{fig:few_shot_redial} (a) and Figure \ref{fig:few_shot_redial} (b), and the decrease of DIST-4 when we increase the training data size.

\subsection{Human Evaluation}
We conduct a human evaluation on Wizard of Wikipedia to assess the overall quality of the responses of our model compared to T5 and RAG\footnote{We use the published FiD RAG DPR model at https://parl.ai/projects/hallucination/}. Specifically, we randomly select 100 responses for each model with the same context from Test Seen and Test Unseen. For the few-shot setting, we use the models trained with 50 dialogues. We hire workers on Amazon Mechanical Turk to rate models' responses on a 0 - 2 scale with three metrics: Fluency, Coherence, and Knowledge. The order of the systems shown to workers is shuffled to avoid confounding practice effects. Three different workers evaluate each dialogue turn. Table \ref{tab:result_human_eval} reports average metrics scores. We observe that responses from our fully-supervised model are more fluent and coherent than those from RAG, which benefits from our simple but effective essential knowledge representation. We can also see significant improvement on all metrics for \modelname\ under a zero-shot setting compared to the T5 model. Performance improvement under the few-shot setting is less than in the zero-shot setting, but \modelname\ still outperforms T5 on all metrics, which aligns with the result in automatic evaluation. Interestingly, we observe that responses from the model trained with 50 dialogues have already been very fluent and coherent, which is even higher than those from the fully-supervised model. However, responses from the fully-supervised model contain the most appropriate knowledge, which suggests that the model has learned how to generate high-quality responses in a few-shot setting, but it continues to learn how to ground on knowledge with more training samples.

\begin{table}[htb!]
    \centering
    \small
    \begin{tabular}{lccc}
    Model & Fluency & Coherence & Knowledge\\
    \midrule
    \midrule
    RAG & 1.06 & 1.08 & 1.19 \\
    \midrule
    \multicolumn{4}{l}{\textbf{T5-Large}}\\
     - Zero-shot & 0.87 & 0.98 & 0.98 \\
     - Few-shot & 1.26 & 1.35 & 1.31\\
    \midrule
    \multicolumn{4}{l}{\textbf{\modelname}}\\
     - Zero-shot & $1.20^{**}$ & $1.34^{**}$ & $1.25^{**}$ \\
     - Few-shot & $\textbf{1.29}^{*}$ & $\textbf{1.42}^{*}$ & $1.39^{**}$ \\
     - Fully-supervised & $1.24^{**}$ & $1.37^{**}$ & $\textbf{1.46}^{**}$ \\
    \end{tabular}
    \caption{Human evaluation results of different models on Wizard of Wikipedia. We test T5 baselines and RAG model against \modelname\ with **p < 0.01, *p < 0.05.}
    \label{tab:result_human_eval}
    \vspace{-3mm}
\end{table}

\subsection{Discussion and Analysis}
\label{discussion}
\begin{figure*}
    \centering
    \includegraphics[trim={0 0 0 0},clip,width=16cm]{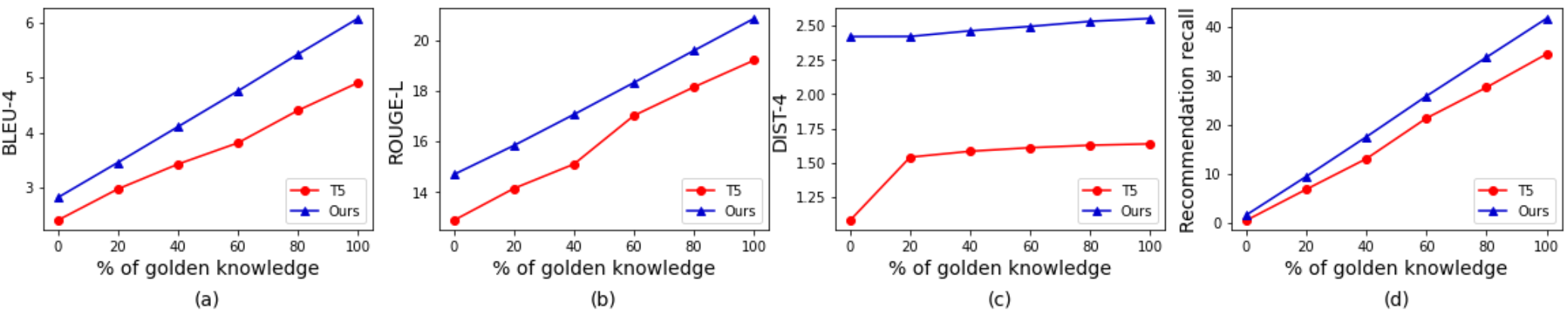}
    \caption{Analysis of models with different knowledge sources on \textsc{ReDial}.}.
    \label{fig:retriever_analysis}
    \vspace{-3mm}
\end{figure*}

To investigate the enormous performance gap between models with golden knowledge and retrieved knowledge in Table \ref{tab:result_redial}, we compare the performance of models with different knowledge sources on the \textsc{ReDial} dataset. Specifically, we mix the golden movies information and the retrieved movie information retrieved in the training/validation/test set to simulate knowledge sources with different recall performances. We experiment with 0/20/40/60/80/100 percent of the golden knowledge. 0 means all training samples includes retrieved knowledge (a flawed knowledge source), 100 means all training samples include golden knowledge (a perfect knowledge source). To have a more realistic setting, we compare the performance of \modelname\ and T5 under the few-shot setting (trained on 50 dialogues), as shown in Figure \ref{fig:retriever_analysis}.

We find that the performance gain for both models is linear with respect to the performance of the knowledge source, whereas \modelname\ has a better boost on the BLEU-4 score and recommendation recall score. The curve with a higher slope shows the potential benefit from our pre-training method when better knowledge sources are available in the future. Furthermore, the gap on DIST-4 between \modelname\ and T5 is almost the same as golden knowledge increases, but the DIST-4 of T5 surprisingly drops when no golden knowledge is available. It means that T5 requires a better knowledge source in the training set to generate diverse responses under a few-shot setting, while \modelname\ has learned that ability in the pre-training process and generates diverse responses out-of-the-box. We also note that the performance boost with a better knowledge source is much more than the generation technology in previous work. This massive gap may shed light on the research direction of knowledge-grounded dialogue tasks for future efforts.

\section{Related Work}
Knowledge-grounded dialogue is becoming an increasingly important topic, with datasets proposed to model its occurrence on different tasks. Dialogues in these datasets are based on various formats of knowledge, such as documents in open-domain conversations \cite{ghazvininejad2018knowledge, dinan2018wizard, gopalakrishnan2019topical}, movie database in movie recommendation conversations \cite{li2018conversational, hayati-etal-2020-inspired}, or knowledge graph in recommendation conversations\cite{moon2019opendialkg, liu2021durecdial}.

One of the principal challenges in knowledge-grounded conversations is incorporating knowledge into dialogue systems. Recent work investigates different techniques of learning a better knowledge representation to fuse knowledge in the response generation process. \citet{ghazvininejad2018knowledge} separately encoded the dialogue history and documents to infuse the response with external world facts. \citet{chen2019knowledgebased, wang2021finetuning, zhou2020improving} joined a knowledge graph representation in a response generation module. \citet{zhu2017flexible} combined the knowledge from the database with the user intent and fed it into the decoder. Unlike these studies, we use a single encoder for both dialogue context and knowledge.

In order to improve the systems' performance on unseen topics and train knowledge-grounded dialogue in a low-resource setting, researchers investigate pre-training methods for the knowledge-grounded tasks. \citet{zhao2020low} pre-trained the dialogue generation model with ungrounded dialogues and the knowledge encoder with the Wikipedia dump separately. \citet{li2021zeroresource} proposed a pre-trained latent variable model to learn the way that the knowledge is expressed in the response. \citet{liu2021three} built a document encoder and a dialogue context encoder, then pre-trained them separately in multiple stages. The knowledge encoder in these studies is pre-trained separately and only accepts the same knowledge format, while we pre-train our model with essential knowledge in the text format, thus fitting different knowledge sources in the downstream tasks. \citet{madotto2020adapter} independently trained adaptors \cite{houlsby2019parameter} for different types of knowledge. In comparison, we use a unified essential knowledge representation in our model. \citet{zhao2020knowledge} and \citet{guu2020realm} pre-trained language models with knowledge selection modules but only focused on document-based generation, limiting their models to document-based knowledge sources.

Inspired by the success of pre-trained language models for a variety of natural language processing tasks \cite{devlin2018bert, radford2019language, yang2019xlnet, ma2021open}, another line of work investigates learning knowledge through language models' parameters \cite{petroni2019language, rosset2020knowledgeaware, roberts2020much}. In our pre-training process, we aim to learn extra knowledge and, more importantly, learn how to generate response grounding on the essential knowledge.

Two recent studies are most closely related to our work. \citet{chen2020kgpt} proposed a pre-trained model for data to text tasks. They unified the knowledge format in the pre-training data and downstream tasks, however only depend on the graph structure and do not work on knowledge-grounded dialogues. \citet{shuster2021retrieval} applied the document retrieval augmentation method \cite{lewis2020retrieval} on open-domain knowledge-grounded dialogues. However, they do not do pre-training and rely on Wikipedia documents in the decoder, limiting their model to document-based dialogues. We use unified essential knowledge instead of documents in our pre-training, making our model more generalizable. Our approach can be seen as generalizing both lines of work, and showing for the first time that a pre-trained model is effective for various knowledge-grounded tasks with different knowledge formats.

\section{Conclusion and Future Work}
We present a knowledge-grounded pre-trained language model \modelname\ that can be applied to various knowledge-grounded dialogue tasks. It subsumes different knowledge sources into a simple but effective unified essential knowledge representation. Evaluation results on two benchmarks indicate that our model performs better in zero-shot and few-shot settings and can generalize to different knowledge grounded tasks.

As future work, we would like to augment our pre-training datasets with more knowledge sources, and apply our method to other knowledge-grounded tasks such as question answering. Another interesting direction would be to develop better information retrievers since experiments show that the retriever is the main bottleneck in the knowledge-grounded dialogues.

\bibliography{anthology,custom}
\bibliographystyle{acl_natbib}

\newpage
\appendix
\section{Implementation Details}
\label{sec:implementation_details}
We process the Reddit monthly submissions and comments dump from 2011 to 2017, consisting of over 894k knowledge-grounded dialogue turns. As detailed in Section \ref{pre_training_datasets_reddit}, we set the threshold as 0.35 in the semantic ranking. After filtering with our hierarchical information extraction method, over 321k dialogue turns remain. All dialogue turns in the OpenDialKG dataset are used in the pre-training. Each dialogue turn is processed to form a sequence of tokens consisting of three segments: dialogue context, essential knowledge, and response. We keep the top-three triples/keywords as our essential knowledge in pre-training and downstream tasks. \modelname\ is implemented with Huggingface Pytorch Transformers\footnote{https://github.com/huggingface/transformers is licensed under the Apache License 2.0} \cite{wolf2020transformers} and initialized with the 800M-parameter T5 model. We use Adam \cite{kingma2014adam} with weight decay for pre-training. Training examples are truncated to ensure a maximal length of 512. Models are pre-trained on 8 Nvidia V100 GPUs until we observe no progress on validation data or up to 20 epochs. The best configuration of hyper-parameters is selected through cross-validated grid-search.

\section{Ethical Considerations}
It is essential to consider potential ethical issues in knowledge-grounded dialogue systems. In our work, \modelname\ is pre-trained on a large-scale dataset Reddit Conversation, which is crawled from the internet. We follow \citet{galley2018end} to filter out dialogues that have profanity content. However, it is still possible to include inappropriate content in the pre-training dataset. In processing the Reddit Conversation dataset during pre-training, we have carefully designed rules to remove knowledge that has profanity words. Additionally, the T5 model may have seen inappropriate content in its pre-training tasks, and it may generate wrong responses even if we input appropriate knowledge. Considerable additional work is needed to detect profanity content when we generate with a pre-trained language model. In addition to these ethical considerations, we have sought to better conduct our human evaluation by transparently communicating with crowd-workers about data use and study intent and compensating workers at a reasonable hourly wage.

\section{Human Evaluation Interface}
\begin{figure*}
    \centering
    \includegraphics[trim={0 0 0 0},clip,width=16cm]{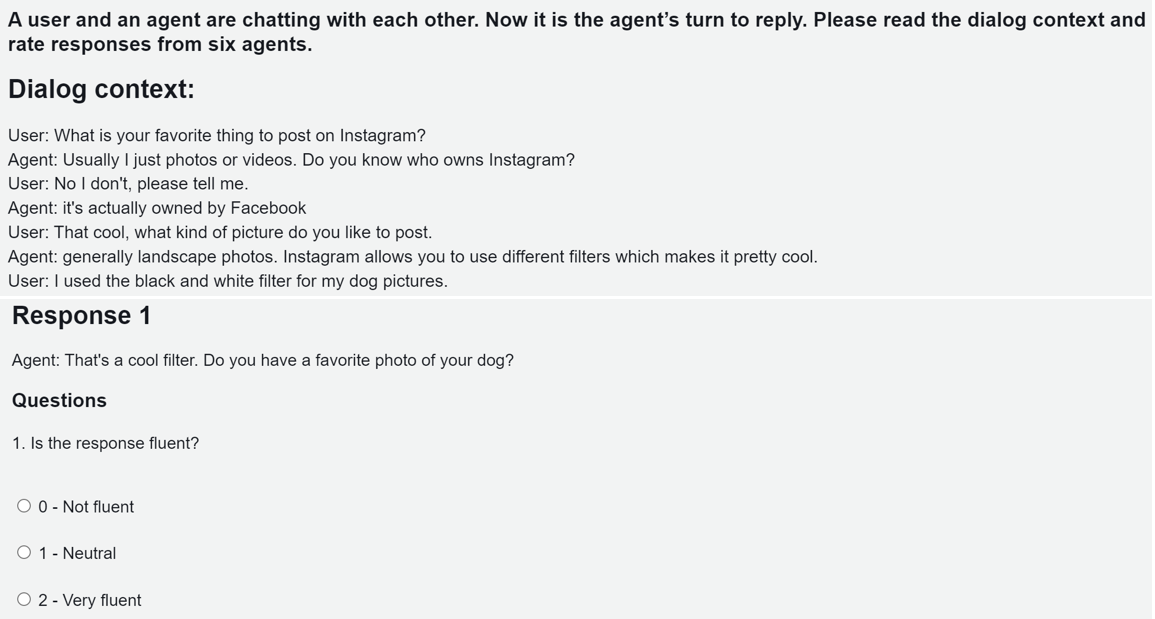}
    \caption{Screenshot of human evaluation interface.}.
    \label{fig:human_eval}
    \vspace{-3mm}
\end{figure*}
Figure \ref{fig:human_eval} shows the interface of an example in our human evaluation.


\end{document}